\documentclass[journal]{IEEEtai}

\usepackage[colorlinks,urlcolor=blue,linkcolor=blue,citecolor=blue]{hyperref}

\usepackage{color}
\usepackage{amsmath,amsfonts}
\usepackage{array}
\usepackage{subfig}
\usepackage{textcomp}
\usepackage{stfloats}
\usepackage{url}
\usepackage{verbatim}
\usepackage{graphicx}
\usepackage{cite}
\usepackage{booktabs}
\usepackage{multirow}
\usepackage{diagbox}
\usepackage{threeparttable}
\usepackage{authblk}

\begin{document}



\title{Diff-Oracle: Deciphering Oracle Bone Scripts with Controllable Diffusion Model} 

\author{Jing Li, Qiu-Feng Wang, \IEEEmembership{Member, IEEE}, Siyuan Wang, Rui Zhang, Kaizhu Huang, \IEEEmembership{Senior Member, IEEE}, and Erik Cambria, \IEEEmembership{Fellow, IEEE}}

\markboth{Journal of IEEE Transactions on Artificial Intelligence, Vol. 00, No. 0, Month 2020}
{First A. Author \MakeLowercase{\textit{et al.}}: Bare Demo of IEEEtai.cls for IEEE Journals of IEEE Transactions on Artificial Intelligence}

\maketitle

\begin{abstract}
Deciphering oracle bone scripts plays an important role in Chinese archaeology and philology. However, a significant challenge remains due to the scarcity of oracle character images. To overcome this issue, we propose Diff-Oracle, a novel approach based on diffusion models to generate a diverse range of controllable oracle characters. Unlike traditional diffusion models that operate primarily on text prompts, Diff-Oracle incorporates a style encoder that utilizes style reference images to control the generation style. This encoder extracts style prompts from existing oracle character images, where style details are converted into a text embedding format via a pretrained language-vision model. On the other hand, a content encoder is integrated within Diff-Oracle to capture specific content details from content reference images, ensuring that the generated characters accurately represent the intended glyphs. To effectively train Diff-Oracle, we pre-generate pixel-level paired oracle character images (i.e., style and content images) by an image-to-image translation model. Extensive qualitative and quantitative experiments are conducted on datasets Oracle-241 and OBC306. While significantly surpassing present generative methods in terms of image generation, Diff-Oracle substantially benefits downstream oracle character recognition, outperforming all existing SOTAs by a large margin. In particular, on the challenging OBC306 dataset, Diff-Oracle leads to an accuracy gain of 7.70\% in the zero-shot setting and is able to recognize unseen oracle character images with the accuracy of 84.62\%, achieving a new benchmark for deciphering oracle bone scripts.
\end{abstract}

\begin{IEEEImpStatement}
The oracle bone script is the earliest known mature writing system in China. The study of this script not only deepens the understanding of ancient Chinese history and culture but also plays a crucial role in uncovering the development patterns of human civilization. However, deciphering oracle characters is challenging due to the scarcity of available data. To overcome this issue, we propose Diff-Oracle, a novel generative framework based on diffusion models designed for generating controllable oracle character images. In terms of character generation, our Diff-Oracle outperforms existing generative methods. Moreover, with the supplementation of images generated by Diff-Oracle and the adoption of a mixup strategy, the recognition accuracy of Oracle characters is significantly improved to a new state-of-the-art benchmark. Our work advances artificial intelligence and machine learning applications while providing invaluable tools for historians and linguists studying ancient civilizations.
\end{IEEEImpStatement}


\section{Introduction}\label{sec:intro}

\begin{figure}[htbp]
\centering
\includegraphics[width=0.49\textwidth]{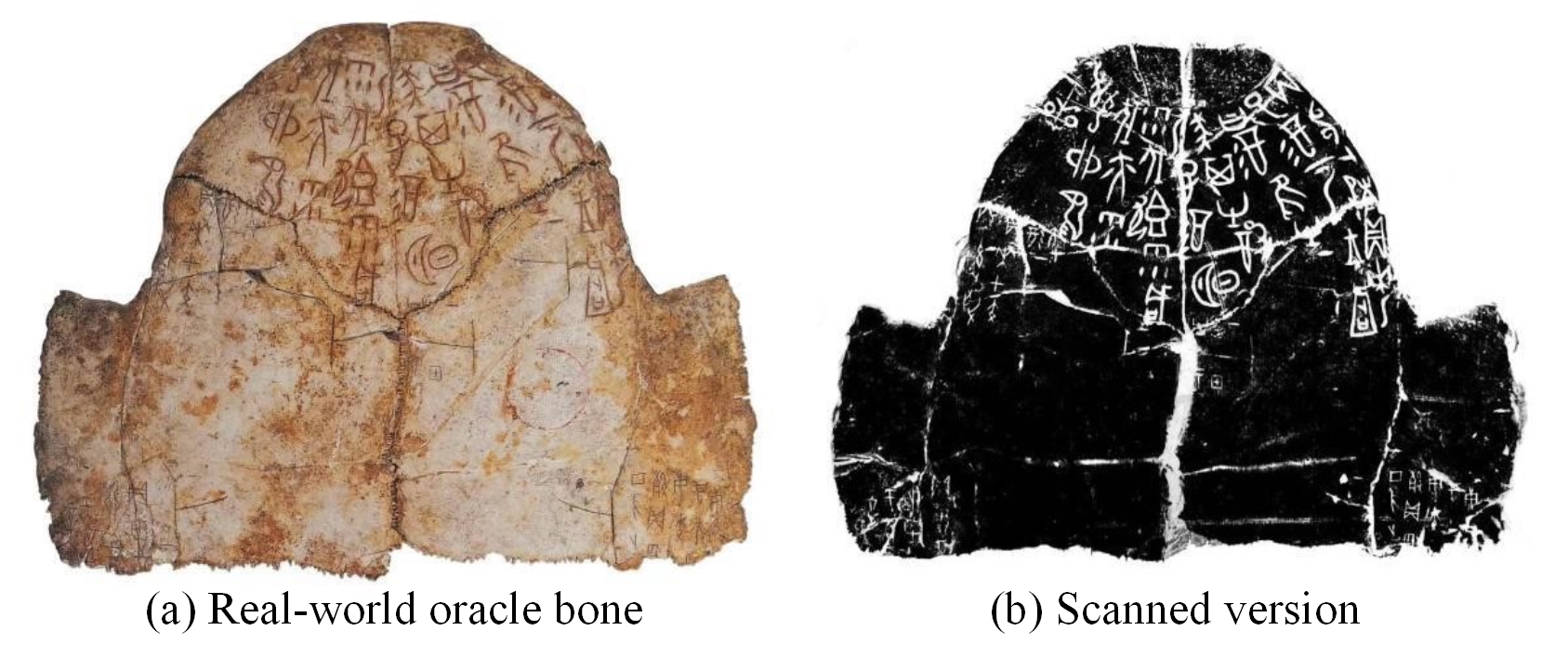}
\caption{Example of an oracle bone (a) and its related scanned rubbings (b).}
\label{fig:intro}
\end{figure}

\IEEEPARstart{T}{he} oracle bone script is the earliest known mature writing system in China, dating back about 3,500 years. It was usually inscribed on turtle nails or animal bones by the rulers of the Shang dynasty. The study of this script is significant for understanding the history of Chinese writing and culture. Fig.~\ref{fig:intro}a shows a piece of oracle bone with characters. Most ancient bones are preserved as rubbings and then scanned into images, as shown in Fig.~\ref{fig:intro}b.

\begin{figure*}[htbp]
\centering
\includegraphics[width=\textwidth]{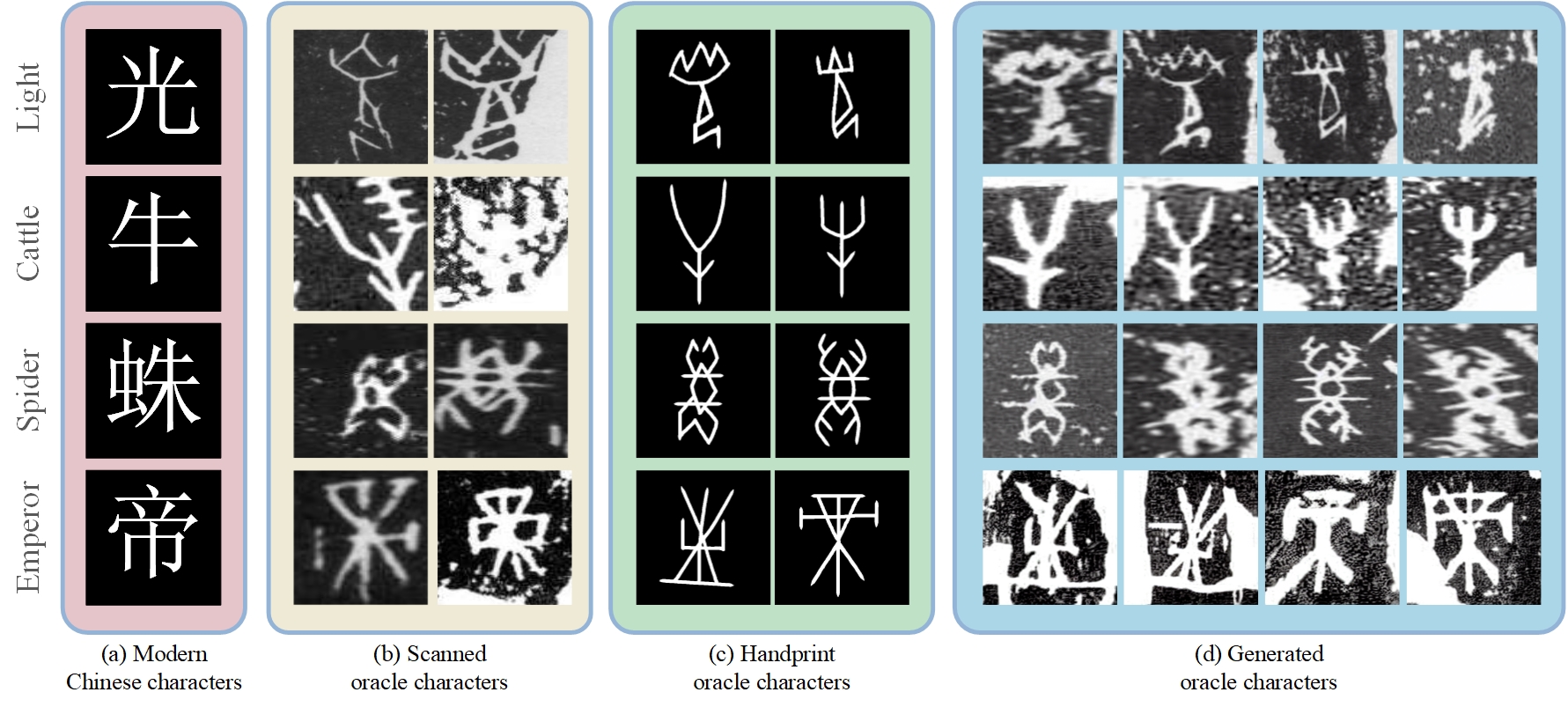}
\caption{Modern Chinese characters (a) and their corresponding scanned oracle characters (b). Given a scanned oracle character image (b) as reference style and a handprinted image (c) as reference content, Diff-Oracle is able to generate realistic and controllable samples (d). Images in the same row belong to the same class.}
\label{fig:intro2}
\end{figure*}

To decipher oracle bone scripts, researchers have focused on isolated characters cropped from scanned rubbings, as shown in Fig.~\ref{fig:intro2}b. These scanned characters are often of poor quality due to long-term burial or improper excavation, resulting in partially missing parts, dense white regions, and bone fractures~\cite{OBC306}. Additionally, due to the lack of uniform writing standards over a long historical period, an oracle character class often shares more than one glyph, leading to severe intra-class variation~\cite{OBC306}.

\begin{figure}[htbp]
\centering
\subfloat[Training set]{
\includegraphics[width=0.48\textwidth]{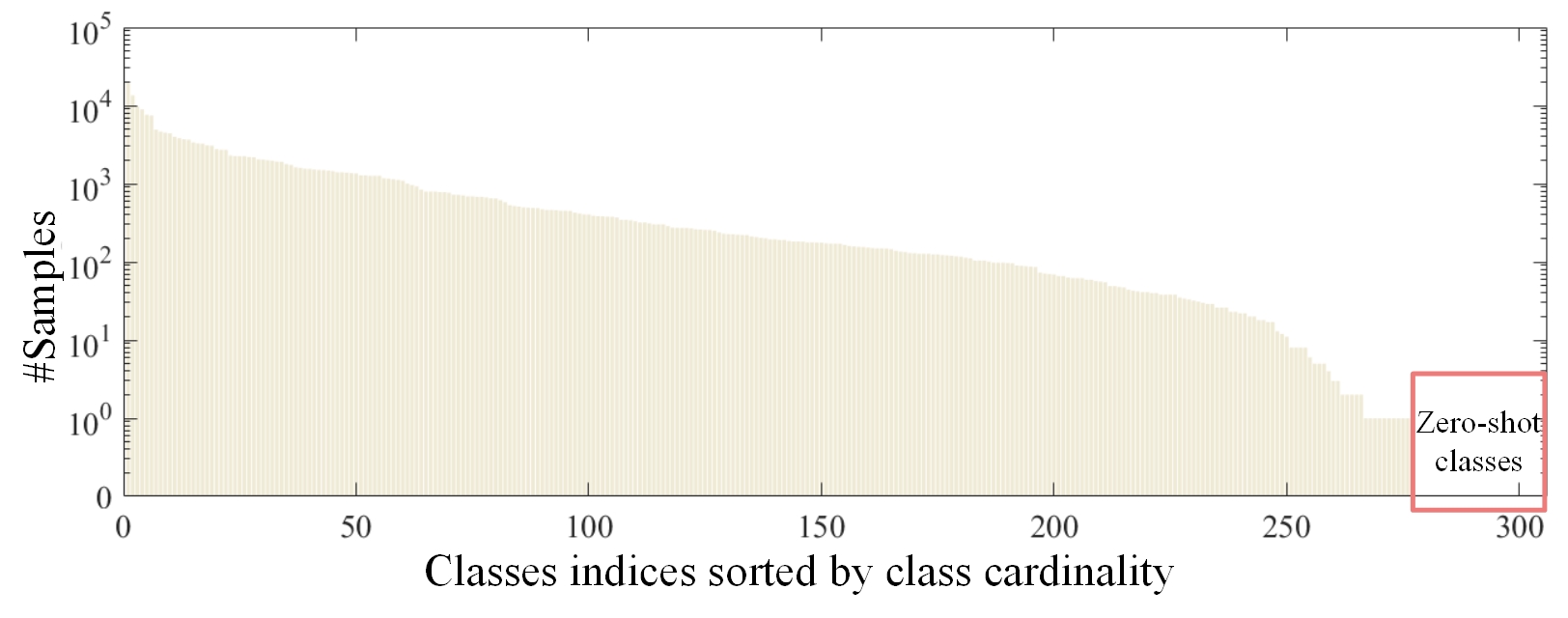}
\label{fig:training_dis}}
\vfill
\subfloat[Test set]{
\includegraphics[width=0.48\textwidth]{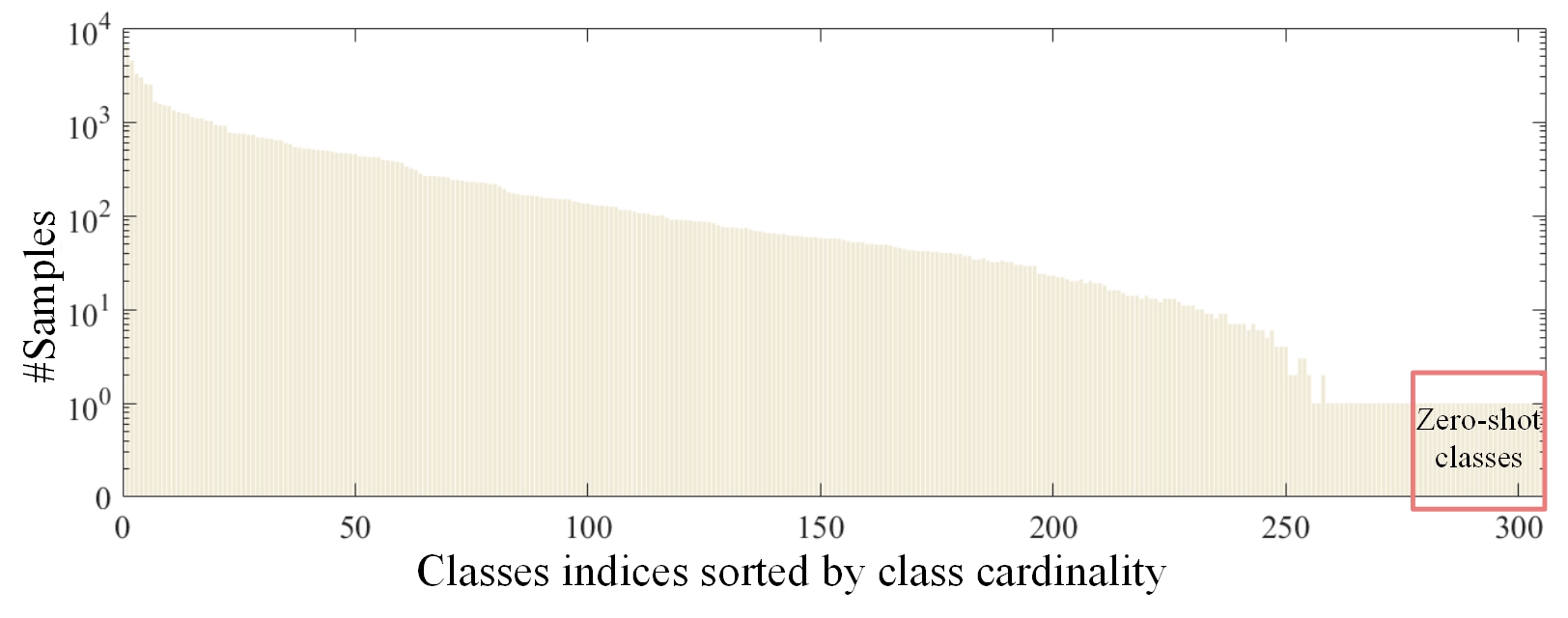}
\label{fig:test_dis}}
\caption{Data distributions of training and test sets from OBC306~\cite{OBC306} sorted by class cardinality.}
\label{fig:obc_dis}
\end{figure}

In contrast to works~\cite{oracleinterpretation,oraclecascadeinterpretation} that focus on the interpretation of oracle characters by translating them into modern Chinese characters, we concentrate on classifying them for recognition. While manual recognition typically requires extensive expertise and high costs, researchers have turned to exploring machine learning~\cite{lv2010graphic,li2011isomorphism,gu2016topology,liu2017svm} and deep learning techniques~\cite{densenet,inceptionv4,DeepLearningBooK} to automatically recognize oracle characters. Despite advances in deep learning, the recognition accuracy for this task remains unsatisfactory~\cite{oracle20k,icdar2019,li2023towards}. One main challenge lies in the scarcity of labeled oracle character data. In particular, unlike modern Chinese characters, collecting images of oracle characters is difficult. Meanwhile, only a very limited number of experts are able to recognize and annotate them, making the process laborious and time-consuming. 
Fig.~\ref{fig:obc_dis} illustrates the distributions of training and test data from a public benchmark data set, OBC306~\cite{OBC306}. It is evident that many classes have insufficient data, and some classes even lack training samples at all when there are test instances (i.e., zero-shot classes in Fig.~\ref{fig:obc_dis}).
To overcome this, we propose a diffusion model tailored to the challenging oracle character generation and recognition. Named Diff-Oracle, our generative model is able to augment sufficient and controllable realistic oracle characters, especially for the challenging zero-shot classes, as shown in Fig.~\ref{fig:intro2}d.

To alleviate data scarcity in oracle character recognition, data augmentation has been explored in two main directions, direct and indirect approaches~\cite{li2021mix,huang2022acm,li2023towards}. The former~\cite{li2021mix,li2023towards,han2020accv,zhao2022accv} generates new data for minority or insufficient classes within the same domain via GANs~\cite{cgan,gamo} or mixup strategies~\cite{mixup,remix,patchup}. However, they are limited by single-domain data and are not applicable to zero-shot classes. The latter~\cite{wang2022tip,huang2022acm} also adopt GAN models but rely on handprinted oracle characters to expand the writing information. Handprinted images are transcribed by experts with high resolution and clean backgrounds (e.g., Fig.~\ref{fig:intro2}c)\footnote{Correspondingly, we call original oracle characters from scanned rubbings (e.g., Fig.~\ref{fig:intro2}b) as scanned images.}, which are easier to obtain. Although these methods enrich scanned data with the help of handprinted data and generate sufficient oracle character images, they often suffer from a lack of controllability and training instability~\cite{karras2020gandata,rangwani2022ganlongtail}. Compared to GAN models~\cite{cgan,acgan,cgannature}, text-to-image diffusion models (DMs), such as Stable Diffusion~\cite{stablediffusion} and ControlNet~\cite{controlnet}, have recently shown greater potential for synthesizing high-quality images~\cite{2023GLIGEN,textinversion,stablediffusion,controlnet}. This is attributed to their stable learning objectives and controllable text prompts. Nevertheless, these models struggle to generate oracle characters with specific styles and contents. The primary obstacles here are twofold: 1) the absence of text prompts for \textbf{styles}, and 2) the lack of \textbf{content} information.

In general, text-to-image DMs typically utilize text prompts to guide resulting image styles, including factors like colors and appearance. However, it is a challenge to use natural language to accurately describe the styles of oracle characters, such as background texture, noise, and stroke thickness, even for experts. As a result, text prompts for DMs in generating oracle characters are unfortunately unavailable. However, this style information is inherently embedded in the oracle characters themselves. This raises the question: \textit{Is it possible to leverage existing oracle character images to control styles required in DMs for oracle character generation?} Inspired by InST~\cite{inst}, we design a style encoder to learn style representations directly from scanned images. Concretely, a pretrained language-vision model, CLIP~\cite{clip}, is adopted to transform visual representations into text embeddings as style prompts, ultimately enabling precise control over the desired style in generating oracle characters.

In addition to style prompts, content information (e.g., character glyphs) is also essential for DMs, which guarantees the desired content of generation (e.g., Diff-Font~\cite{diff-font} and ControlNet~\cite{controlnet}). In the field of general font or image generation, current wisdom~\cite{diff-font,controlnet,2023GLIGEN} often leverages label information or sketch information as contents. However, the significant intra-class variance in oracle characters complicates the task of summarizing their content with mere labels. 
Furthermore, extracting accurate sketch information (e.g., Canny edges) from low-quality scanned images of oracle bones (e.g., the degraded images shown in Fig.~\ref{fig:intro2}b) is nearly impossible. Meanwhile, handprinted oracle characters (e.g., Fig.~\ref{fig:intro2}c) readily offer such content information, yet lack pixel-level alignment with scanned images necessary for DM training. It is noted that handprinted and scanned oracle characters are paired at the class level in the Oracle-241 dataset~\cite{wang2022tip}, while they do not match at the pixel level (e.g., comparing Fig.~\ref{fig:intro2}b and Fig.~\ref{fig:intro2}c). Therefore, \textit{it is crucial to obtain the pixel-level paired oracle character images for effective content control during DM training}. To address this, we additionally train an image-to-image translation model, CUT~\cite{cut}, to convert scanned characters into pseudo handprinted ones (e.g., Fig.~\ref{fig:cut}), forming pixel-level paired data. During training, these pseudo handprinted characters serve as content images. To maximize their utility, motivated by ControlNet~\cite{controlnet}, we propose a content encoder to extract glyph information from these images and integrate it into Diff-Oracle. 

Our main contributions are summarized as follows:
\begin{enumerate}
  \item We propose Diff-Oracle, a novel diffusion model tailored to the challenging oracle character generation, which is able to generate a diverse range of controllable oracle characters. To our knowledge, we are the first to explore DMs for both oracle character generation and recognition.
  \item We propose a style encoder to extract style prompts directly from available oracle character images, eliminating the need for text prompts for generation control.
  \item We propose a content encoder to extract glyph information for oracle characters. To effectively train Diff-Oracle, we generate pseudo handprinted characters as content images, ensuring pixel-level pairing with scanned oracle characters. 
  \item We propose a two-stage training strategy to enhance the optimization of Diff-Oracle in terms of style and content control. 
  \item Through quantitative and qualitative experiments, we validate the effectiveness of Diff-Oracle for oracle character generation. Augmenting the training data with images generated by Diff-Oracle, along with a mixup strategy, we improve the recognition accuracy on Oracle-241 and OBC306, surpassing existing state-of-the-art (SOTA) recognition methods. In particular, Diff-Oracle achieves an impressive accuracy of 84.62\% in zero-shot classes on OBC306, setting a new SOTA standard for deciphering oracle bone scripts.
\end{enumerate}

\section{Related Works}
\label{sec:relat}

\subsection{Oracle Character Generation}
We briefly introduce two main categories of oracle character generation methods: direct and indirect. Direct methods~\cite{li2021mix,li2023towards,han2020accv,zhao2022accv} focuses on generating data within the same domain. For example, various works~\cite{li2021mix,li2023towards} generated new features for minority classes in the latent space via mixup strategies and GANs, ultimately improving the classification accuracy of minority oracle character classes. Furthermore, Han et al.~\cite{han2020accv} and Zhao et al.~\cite{zhao2022accv} focused on few-shot learning of handprinted data and augmented few-shot classes at the pixel level by transforming the stroke vectors of characters. In contrast, indirect methods synthesize data by translating oracle characters from the handprinted domain to the scanned domain. For instance, STSN~\cite{wang2022tip} decomposed oracle characters into structure and texture features, then scanned data can be synthesized by a combination of handprinted structure features and scanned texture features. Similarly, AGTGAN~\cite{huang2022acm} translated handprinted data to scanned data by transforming glyphs and transferring textures. However, most of these approaches are based on GANs, which face challenges in meeting specific requirements for detailed controllability. In addition, without careful training, they probably suffer from mode collapse, unstable training, and pixel artifacts~\cite{karras2020gandata,rangwani2022ganlongtail}.

\subsection{Image-to-Image Translation}\label{related_translation}
Image-to-image translation involves translating an image from one domain to another while preserving its essential content and structure. GAN-based approaches~\cite{pix2pix,cyclegan,cut} have dominated this field for many years. Pix2Pix~\cite{pix2pix} employed a conditional GAN~\cite{cgan} to produce clear transformed images from reference images. Subsequently, CycleGAN~\cite{cyclegan} eliminated the need for paired data by introducing a cycle-consistency loss. CUT~\cite{cut} further simplified the process by removing bidirectional translation from CycleGAN and incorporating contrastive learning. Inspired by CUT and CycleGAN, many follow-up approaches have emerged~\cite{cyclefamily1,cyclefamily2,cut3,cut4}. Recently, a large number of DM-based methods have been introduced in this field and achieved promising results~\cite{unitddpm,egsde,Palette,vqbb,diffphoto}. Nevertheless, most DM models use text prompts to control style, which is unfortunately infeasible for oracle characters.

\subsection{Chinese Font Generation}\label{related_font}
Chinese font generation, a subtask within image-to-image translation, focuses on changing font style while retaining the original semantic content. Early works relied on classic image generation approaches with encoder-decoder architectures~\cite{tian2017zi2zi,2017aegn}.
However, the complexity of font images required more customized models to effectively capture their unique characteristics. Subsequent studies incorporated font-specific prior information~\cite{2021dgfont,2022fsfont,2023cffont} and additional data~\cite{2019scfont,2020dmfont} to improve design. Nevertheless, Chinese font generation models may not be directly applicable to oracle character generation. The primary challenge lies in the low quality of oracle character images, which typically contain severe noise. As a result, it is difficult to extract essential sketch information, such as strokes.

\subsection{Diffusion Models}\label{related_diff}
Currently, DMs have been widely utilized in image generation tasks~\cite{classifierfree,classifierguidance}, demonstrating excellent fidelity and diversity. The concept of DMs was first introduced in~\cite{nonequilibrium} and promoted by DDPM~\cite{ddpm}. 
To reduce computation costs and improve the accessibility of these powerful DMs, Latent Diffusion Models (LDMs)~\cite{stablediffusion} were introduced, moving the diffusion process to the latent space. Among them, Stable Diffusion (SD) is a successor of LDMs conditioned on text embeddings from the CLIP text encoder~\cite{clip}. Personalized frameworks such as GLIGEN~\cite{2023GLIGEN} and ControlNet~\cite{controlnet} fine-tuned pretrained SD with additional input conditions. Beyond general research, numerous works have tailored DMs for specific tasks. For example, Textual Inversion~\cite{textinversion} and InST~\cite{inst} leveraged pretrained text-to-image DMs for personalized style control. Moreover, Fill-Up~\cite{fillup} adopted Textual Inversion to augment training data on long-tailed learning. However, applying DMs directly to oracle character generation presents two main challenges: the absence of text prompts limits style control, and the lack of content images hinders precise content control.

\section{Methodology}

\begin{figure*}[htbp]
\centering
\includegraphics[width=\textwidth]{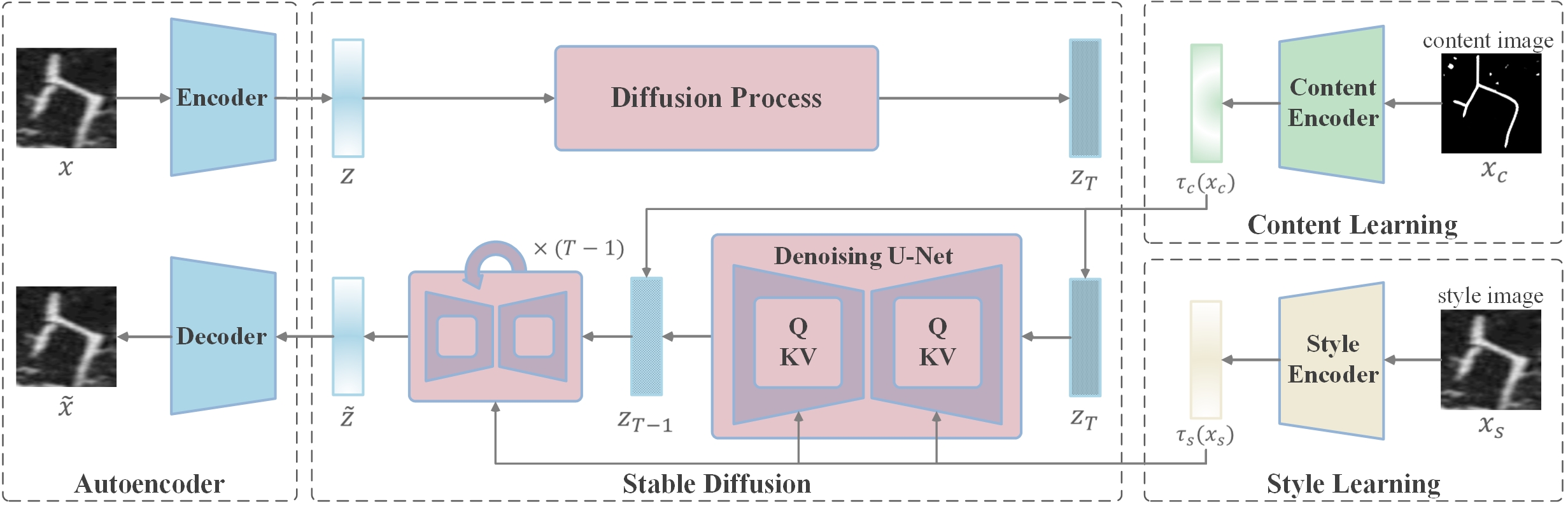}
\caption{Overall architecture of Diff-Oracle including four blocks: Autoencoder, Stable Diffusion, Content Learning, and Style Learning. During the training process, an oracle character image $x$, a pixel-level matched content image $x_c$, and a style image $x_s$ ($x=x_s$ here) are input to the model. Then, style and content information can be extracted from the style encoder $\tau_s$ and the content encoder $\tau_c$, respectively. Meanwhile, Encoder in the Autoencoder block extracts features from $x$, which places the diffusion process in the latent space. Finally, based on these extracted features, the Stable Diffusion block is fine-tuned, and the style encoder $\tau_s$ and the content encoder $\tau_c$ are trained. In the generation phase, given a handprinted image as content $x'_c$ and a scanned image as style $x'_s$, a new oracle character $\tilde{x}$ can be generated by Diff-Oracle from random noise $z_T$, which has the same content as $x'_c$ and the same style as $x'_s$.}
\label{fig:diff_overview}
\end{figure*}

\subsection{Overview}
Diff-Oracle is built upon Stable Diffusion (SD)~\cite{stablediffusion}, which utilizes textual guidance to control the style of generated images. However, it is challenging to describe the desired styles of oracle characters, such as stroke thickness, tips, and joined-up writing patterns.
To overcome this issue, instead of using text prompts, we propose a style encoder $\tau_s$ that directly encodes stylistic features from a given style image $x_s$ and converts them into textual embeddings $\tau_s(x_s)$. 
Subsequently, we observe that using style conditions alone cannot precisely control image content. Inspired by ControlNet~\cite{controlnet}, we incorporate a content encoder $\tau_c$ into our framework to manage content generation, which extracts content features $\tau_c(x_c)$ from a content image $x_c$. 

Fig.~\ref{fig:diff_overview} provides an overview of Diff-Oracle, which consists of a frozen autoencoder, a pretrained SD, a trainable content encoder, and a trainable style encoder. 
During the training process, we feed Diff-Oracle with a scanned oracle character image $x$, a pixel-level matched content image $x_c$ and a style image $x_s$ (here $x_s=x$). Style and content information is then extracted from the style encoder $\tau_s$ and the content encoder $\tau_c$, respectively. At the same time, Encoder in the Autoencoder block extracts features from $x$ to put the diffusion process into the latent space. Finally, based on these extracted features, SD is fine-tuned while the content encoder and the style encoder are trained. In the generation phase, given a handprinted image as the content $x'_c$ and a scanned image as the style $x'_s$, a new scanned oracle character $\tilde{x}$ can be generated with the same content as $x'_c$ and the same style as $x'_s$ by the well-trained Diff-Oracle.

The remainder of this section is organized as follows: we begin by describing the details of the style encoder in Section~\ref{styleencoder} and content encoder in Section~\ref{contentencoder}; next, the overall training strategy is presented in Section~\ref{learningprocedure}; finally, the adopted strategy to achieve guidance with different levels of style and content is introduced in Section~\ref{classifier-free}.

\subsection{Style Guidance}\label{styleencoder}
In the field of text-to-image DMs, most works~\cite{stablediffusion,2023GLIGEN,controlnet} rely heavily on extensive text prompts to accurately define image styles, which control synthesized images with specific desired styles. However, unlike natural image styles, such as color and appearance, oracle character styles are more complicated, also including various stroke thicknesses, tips, and/or joined-up writing patterns like Chinese character images~\cite{2021dgfont}. It is difficult, if not possible, to describe these styles in natural language, even for experts. Inspired by InST~\cite{inst}, we introduce a style encoder into Diff-Oracle. This style encoder is designed to learn style representation directly from an oracle character image, which serves as style prompts to guide the synthesis process towards desired styles.

\begin{figure}[htbp]
\centering
\includegraphics[width=0.49\textwidth]{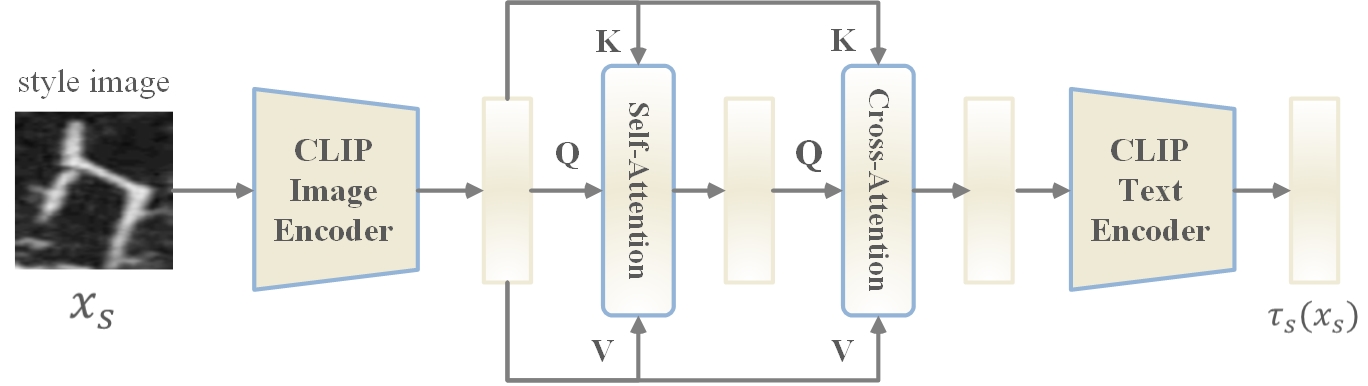}
\caption{Architecture of style encoder $\tau_s$ comprises three modules: CLIP Image Encoder $\tau_{s1}$, Multi-layer Attention $MultiAtt$ and CLIP Text Encoder $\tau_{s2}$. Style input is initially processed by $\tau_{s1}$ to obtain the visual embedding, followed by $MultiAtt$ to emphasize the style information, and ultimately by $\tau_{s2}$ to obtain the style information in text embedding format.}
\label{fig:style_encoder}
\end{figure}

Fig.~\ref{fig:style_encoder} presents the structure of our style encoder. To align with the SD framework, the format of style representation should be consistent with the standard text embedding format for the baseline SD. This standard format is typically generated by the CLIP text encoder when processing natural language inputs. Here, it can be achieved by employing the CLIP image encoder and CLIP text encoder, since CLIP has the intrinsic property of aligning the image embedding and the text embedding in the latent space~\cite{clip}. 
Therefore, as illustrated in Fig.~\ref{fig:style_encoder}, the style encoder $\tau_s$ is composed of three modules: CLIP image encoder $\tau_{s1}$, multi-layer attention $MultiAtt$, and CLIP text encoder $\tau_{s2}$. We first feed the style image $x_s$ into $\tau_{s1}$ to obtain the image embedding $\tau_{s1}(x_s)$. This image embedding is input to attention layers, which emphasize the pivotal style information of $x_s$. 
Each layer implements $\text{Attention}(Q,K,V)=\text{softmax}(\frac{QK^T}{\sqrt{d}})\cdot V$, with:
\begin{equation}
\begin{split}
  Q_i=W_Q^{(i)} \cdot v_i, K=&W_K^{(i)} \cdot \tau_{s1}(x_s), V=W_V^{(i)}\cdot \tau_{s1}(x_s),\\
  v_{i+1}=&\text{Attention}(Q_i,K,V).
\end{split}
\end{equation}
Here, $v_0=\tau_{s1}(x_s)$, $d$ denotes the dimension of features, and we adopt two attention layers. Finally, the desired style representation can be attained by performing $\tau_{s2}$ on $v_2$. 

In summary, the entire process can be simplified as $\tau_s(x_s)=\tau_{s2}(\text{MultiAtt}(\tau_{s1}(x_s)))$. In the style learning phase, we minimize the following optimization objective: 
\begin{equation}
\label{objective_style}
L_{S}=\mathbb{E}_{\epsilon \sim \mathcal N(0,1),\mathcal E(x),t,x_s}\left [ \left \| \epsilon-\epsilon_{\theta}(z_t,t,\tau_s(x_s)) \right \| _2^2\right],
\end{equation}
where $\mathcal E$ denotes the Encoder of Autoencoder, $\epsilon_{\theta}$ denotes the U-Net of SD, and $z_t$ represents the noisy version of latent features extracted from $\mathcal E$ by adding noise for $t$ times.

\subsection{Content Guidance}\label{contentencoder}
In addition to the style conditions, the contents of oracle characters are required to control the specific glyphs of generated characters. 
While class labels can intuitively be the first option for such control, they are not suitable for oracle character recognition. Given the large intra-class variation observed in oracle characters, a single label may be insufficient to describe multiple glyphs in the same class. To address this limitation, we introduce a more fine-grained condition by incorporating handprinted oracle characters. Handprinted images allow for more precise control of the generation process, particularly concerning the specific glyphs and structures desired in the output.

During the training of DMs like ControlNet~\cite{controlnet}, pixel-level pairs of content and input images are necessary; otherwise, they cannot be merged in the model to learn accurate spatial conditioning controls. However, the available handprinted oracle characters are paired with scanned characters at the class level rather than the pixel level. 
To address this limitation, we simply opt for training an off-the-shelf image-to-image translation method, CUT~\cite{cut}, to generate pseudo handprinted images corresponding to scanned images.

CUT is a technique designed to translate an image from one domain to another while preserving its essential content and structure~\cite{cut}. Specifically, CUT consists of a generator and a discriminator, trained in adversarial loss and contrastive loss to achieve one-sided unpaired image-to-image translation. Here, adversarial loss is utilized to encourage the generator to produce images visually similar to target images in competition with the discriminator. Meanwhile, the contrastive loss is formulated to maximize the similarity between corresponding patches of the input source image and the translated target image.

\begin{figure}[htbp]
\centering
\includegraphics[width=0.49\textwidth]{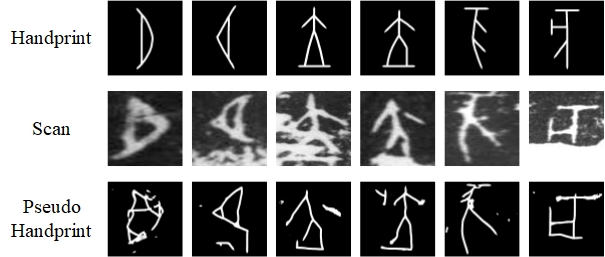}
\caption{Comparison among real handprinted characters (top row), scanned characters (middle row), and pseudo handprinted characters generated by CUT (bottom row).}
\label{fig:cut}
\end{figure}

In our work, CUT is optimized on the class-level paired cross-domain oracle characters, aiming to learn a pixel-level mapping from the scanned to the handprinted domain.
Fig.~\ref{fig:cut} shows examples of generated pseudo handprinted images (the third row). Compared to real handprinted characters (the first row), it is clear that the pseudo handprinted images exhibit a close spatial structural alignment with the glyphs of the scanned images, as depicted in the second row. Therefore, we can pair the pseudo handprinted characters with the scanned characters at the pixel level. Although pseudo handprinted images usually contain some noise, they do not seriously affect those glyphs. Therefore, we currently employ these images generated by CUT to train Diff-Oracle.

After obtaining paired data, we regard a pseudo handprinted image as the content image $x_c$ for training Diff-Oracle. Inspired by ControlNet~\cite{controlnet}, we introduce a content encoder $\tau_c$ to extract the corresponding content features $\tau_c(x_c)$. 
The structure of $\tau_c$ contains eight convolution layers. Each convolution layer includes $3\times3$ kernels and $1\times1$ padding, and is activated by SiLU~\cite{silu}. 
Combined with the style condition, Diff-Oracle can be optimized by minimizing the following objective:
\begin{equation}
\label{objective_content}
\begin{split}
L_{SC}=\mathbb{E}&_{\epsilon \sim \mathcal N(0,1), \mathcal E(x),t,x_s,x_c}\left[ \left \| \epsilon-\epsilon_{\theta}(z_t,t,\tau_s(x_s), \tau_c(x_c)) \right \| _2^2 \right].
\end{split}
\end{equation}
It is worth noting that during the generation stage, we leverage real handprinted characters as content images instead of pseudo ones, since there is no need for pixel-level paired data at this stage, and real handprinted characters are of higher quality in terms of glyph integrity and accuracy.

\subsection{Learning Procedure}\label{learningprocedure}
\begin{figure}[htbp]
\centering
\subfloat[First training stage]{
\includegraphics[height=5cm]{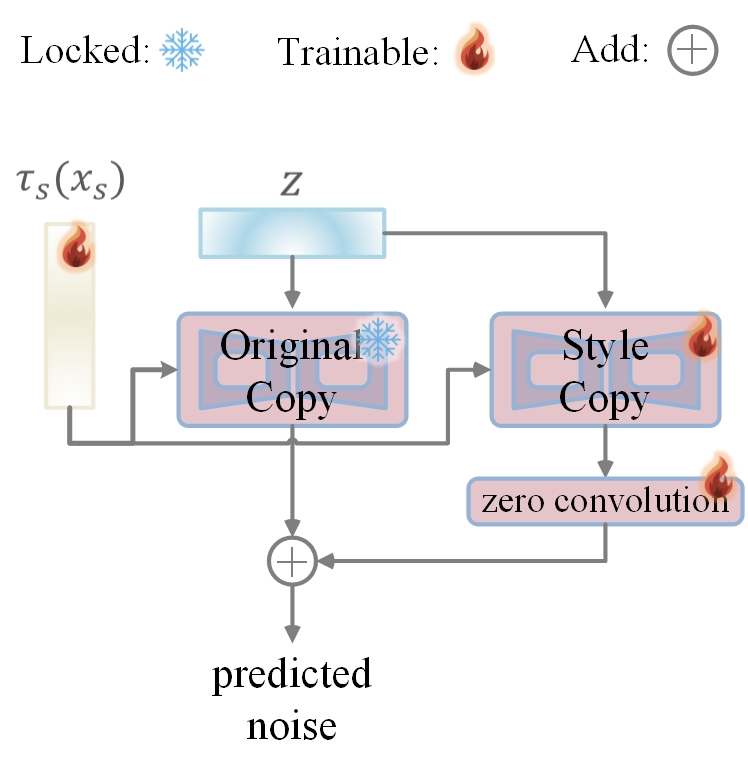}
\label{fig:stage1}}
\vfill
\subfloat[Second training stage]{  \includegraphics[height=5cm]{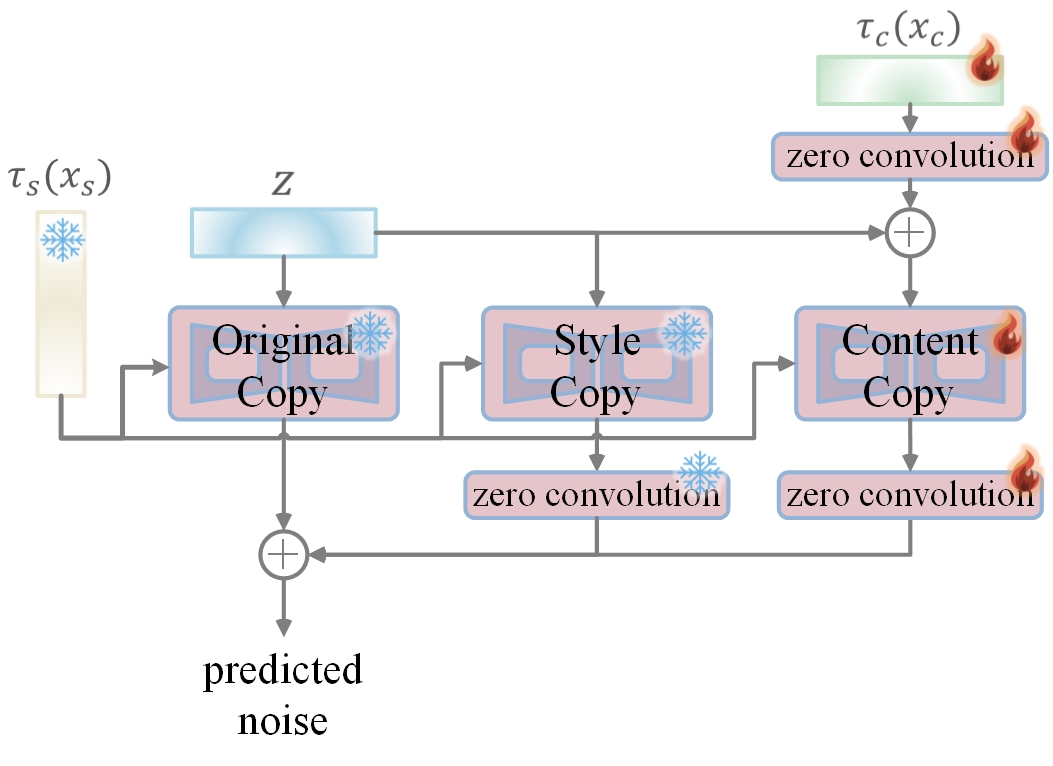}
\label{fig:stage2}}
\caption{Two-stage training strategy of Diff-Oracle. In the first stage, ``Style Copy" and style encoder $\tau_s$ are trained under style condition $x_s$. In the second stage, ``Content Copy" and content encoder $\tau_c$ are trained under content condition $x_c$ while freezing ``Style Copy" and style encoder $\tau_s$.}
\label{fig:controlnet}
\end{figure}

To enhance the impact of style and content controls on the final outputs, we propose a two-stage training strategy. In the first training phase, the style encoder $\tau_s$ and the U-Net of SD $\epsilon_\theta$ are learned according to Eq.~\ref{objective_style}, given a set of conditions including the time step $t$, the input image $x$, and the style image $x_s$. Here, $x$ and $x_s$ are the same scanned images. 
The scale of oracle datasets is small compared to the billions of images normally used to train SD. Thus, directly fine-tuning the pretrained U-Net could lead to overfitting and inferior generalization ability. To overcome this issue, we borrow the key idea introduced in ControlNet~\cite{controlnet} to adjust the DM as illustrated in Fig.~\ref{fig:stage1}. We can see that the parameters of the U-Net inherent to the pretrained SD are duplicated into ``Original Copy" and ``Style Copy". The ``Original Copy" retains the powerful generative ability learned from a vast number of images. At the same time, ``Style Copy" is optimized to learn conditional image generation under the stylistic constraints extracted from scanned oracle characters. These two blocks are connected by a zero convolution layer, where the convolution weights gradually change from zero to appropriate values. This fine-tuning strategy not only preserves production-ready parameters but also augments the robustness of training.

In the second training stage, the emphasis shifts towards training the content encoder $\tau_c$ with well-optimized style information. The extracted content representation $\tau_c(x_c)$ is added to the training. The optimization process follows Eq.~\ref{objective_content}. Here, $x$ and $x_s$ are the same scanned images, and $x_c$ is the corresponding pseudo handprinted image generated via CUT. In order to enable the model to support additional content control, we adopt a fine-tuning strategy similar to that of ControlNet, as shown in Fig.~\ref{fig:stage2}. ``Style Copy" is first locked, and its parameters are then copied to ``Content Copy". This is because ``Style Copy" is personalized to oracle datasets, making it a more suitable starting point than ``Original Copy". Next, the content encoder encodes $x_c$ into feature maps denoted as $\tau_c(x_c)$, which are the same size as $z_t$. Finally, $\tau_c(x_c)$ is added to $z_t$ for denoising learning.

\subsection{Multi-modal Generation}\label{classifier-free}

With a trained Diff-Oracle, we then require an inference strategy that allows Diff-Oracle to generate the final output accurately and diversely. Here, we utilize the concept of classifier-free guidance (CFG)~\cite{classifierfree}. CFG employs a conditional model and an unconditional model, and the final results are obtained by linear extrapolation of the outputs of these two models. Meanwhile, CFG utilizes an adjustable scaling factor to control the balance between fidelity and diversity of the final results. A straightforward application of this strategy for Diff-Oracle is characterized as follows:
\begin{equation}
\begin{split}
  \hat{\epsilon_\theta}&(z_t,t,\tau_s(x_s),\tau_c(x_c))=\epsilon_\theta(z_t,t,0,0)\\
  &+s\cdot (\epsilon_\theta(z_t,t,\tau_s(x_s),\tau_c(x_c))-\epsilon_\theta(z_t,t,0,0)),
\end{split}
\end{equation}
where $s$ represents a scaling factor, and $0$ indicates that this condition is absent for the unconditional model. However, in some cases, styles may be inconsistent with contents due to the discrepancy between the distributions of real and pseudo handprints generated by CUT. In the above formulation, style and content conditions are governed by a single factor, and we are unable to prioritize the importance of one over the other. Therefore, we draw on PAIR Diffusion~\cite{goel2023pair} and introduce the following two scales for content and style:
\begin{equation}
\label{denoise_function}
\begin{split}
  \hat{\epsilon_\theta}&(z_t,t,\tau_s(x_s),\tau_c(x_c)) = \epsilon_\theta(z_t,t,0,0)\\
  &+s_1 \cdot (\epsilon_\theta(z_t,t,0,\tau_c(x_c))-\epsilon_\theta(z_t,t,0,0))\\
  &+s_2 \cdot (\epsilon_\theta(z_t,t,\tau_s(x_s),\tau_c(x_c))-\epsilon_\theta(z_t,t,0,\tau_c(x_c))).
\end{split}
\end{equation}
Here, $s_1$ and $s_2$ represent the scales for the content and style conditions, respectively. It is noted that we never remove content information, as without it, style information becomes meaningless. This way enables us to specify different intensities for style and content conditions during generation. 
To speed up the generation process, we adopt the DDIM algorithm~\cite{ddim} with 50 steps.

\section{Experiments}

\subsection{Datasets}
\label{sec:datasets}
In the experiments, all models are evaluated on \textbf{OBC306}~\cite{OBC306} and \textbf{Oracle-241}~\cite{huang2022acm}. In addition, \textbf{Oracle-AYNU}~\cite{icdar2019} is a handprinted dataset used only as content control for OBC306. More information on the three datasets is detailed in Supplementary Section~\ref{m_dataset}.

\subsection{Implementation Details}
\subsubsection{Network Architecture}
In the recognition task, we adopt a typical 
oracle character recognition work~\cite{li2023towards} in our experiments, which is a ViT-Base model pretrained on ImageNet. Inspired by works~\cite{li2021mix,li2023decouple,li2023towards}, during training, we additionally utilize Repatch, a mixup strategy proposed in the work~\cite{li2023towards}, to fully exploit available oracle data.
\subsubsection{Experimental Setup}
In the generation experiments, we set the image size $256\times 256$ and adopt the AdamW optimizer in two-stage training for Oracle-241 and OBC306. $s_1=2$ and $s_2=5$ for Oracle-241, $s_1=2$ and $s_2=1$ for OBC306. Relevant ablation studies can be seen in Supplementary Section~\ref{scaling_factor}. In the recognition experiments, image size is also $256\times 256$. SGD optimizer is adopted with momentum 0.9 and weight decay 1e-4. More details are described in Supplementary Section~\ref{implementation}.
\subsubsection{Evaluation Metrics}
For generation, to quantitatively compare Diff-Oracle with other generative methods, we adopt two widely-used metrics in the image generation task, i.e., FID~\cite{fid} and LPIPS~\cite{lpips}. We generate 100 instances for each class and calculate the LPIPS distance between pairs of samples, then get the class average as the final LPIPS score. For recognition, following the setting in AGTGAN~\cite{huang2022acm}, for the large-scale OBC306 dataset, we supplement minority classes with generated data to an average number of samples, which is 805 in this paper. For the relatively small Oracle-241 dataset, we supplement each class with generated data to 520 samples. We conduct an ablation study on the number of generated data for Oracle-241 described in Supplementary Section~\ref{number_generated}. As shown in Fig.~\ref{fig:obc_dis}, the test sets of oracle datasets are imbalanced. Therefore, for a fair comparison, we adopt both total and average class accuracy to evaluate the oracle recognition performance following the work~\cite{li2023towards}, which are defined by
\begin{equation}
  total=\frac{1}{N}\sum_{i=1}^{C}r_{i},\text{ } average=\frac{1}{C}\sum_{i=1}^{C}\frac{r_{i}}{n_{i}}.
\end{equation}
Here, $N$ represents the number of test images, $C$ represents the number of classes, $r_i$ denotes the number of correctly classified test images of class $i$, and $n_i$ denotes the total number of test images of class $i$. Following AGTGAN~\cite{huang2022acm}, we also report the total accuracy for zero-shot classes (no training scanned data) in OBC306.

\subsection{Comparison to Previous Methods}
To show the superiority of Diff-Oracle, we conduct comparisons with existing SOTA generative models, including CUT~\cite{cut}, DG-Font~\cite{2021dgfont}, STSN~\cite{wang2022tip}, AGTGAN~\cite{huang2022acm}, InST~\cite{inst}, and ControlNet~\cite{controlnet}. For fair comparisons, we re-trained all models on the same oracle character datasets. In the ControlNet, due to the lack of pixel-level paired data, we adopt existing class-level paired handprinted data as content images in the fine-tuning process. In addition, given the absence of text prompts for oracle datasets, we simply set a uniform prompt of ``Gray scale image of oracle character on the rubbing" for each image.
During the generation process, we provide the corresponding real handprinted data to CUT, DG-Font, STSN, AGTGAN, and ControlNet based on specified classes for generating scanned data. Both InST and our Diff-Oracle require handprinted and scanned data. Hence, we randomly select handprinted and scanned data as input pairs from the same class. As for zero-shot classes in OBC306, we randomly choose scanned data from other classes as style references. 

Meanwhile, we demonstrate the effectiveness of our generated characters in the recognition task. The recognition model is trained on the combination of real data and generated data from different generative models, including our Diff-Oracle, CUT~\cite{cut}, DG-Font~\cite{2021dgfont}, STSN~\cite{wang2022tip}, AGTGAN~\cite{huang2022acm}, InST~\cite{inst}, and ControlNet~\cite{controlnet}. In addition to comparing Diff-Oracle with the above generative models, we also evaluate it with the current SOTA oracle character recognition methods~\cite{icdar2019,li2021mix,li2023towards}.

\subsubsection{Visual Comparison}
\begin{figure*}[ht]
\centering
\includegraphics[width=\textwidth]{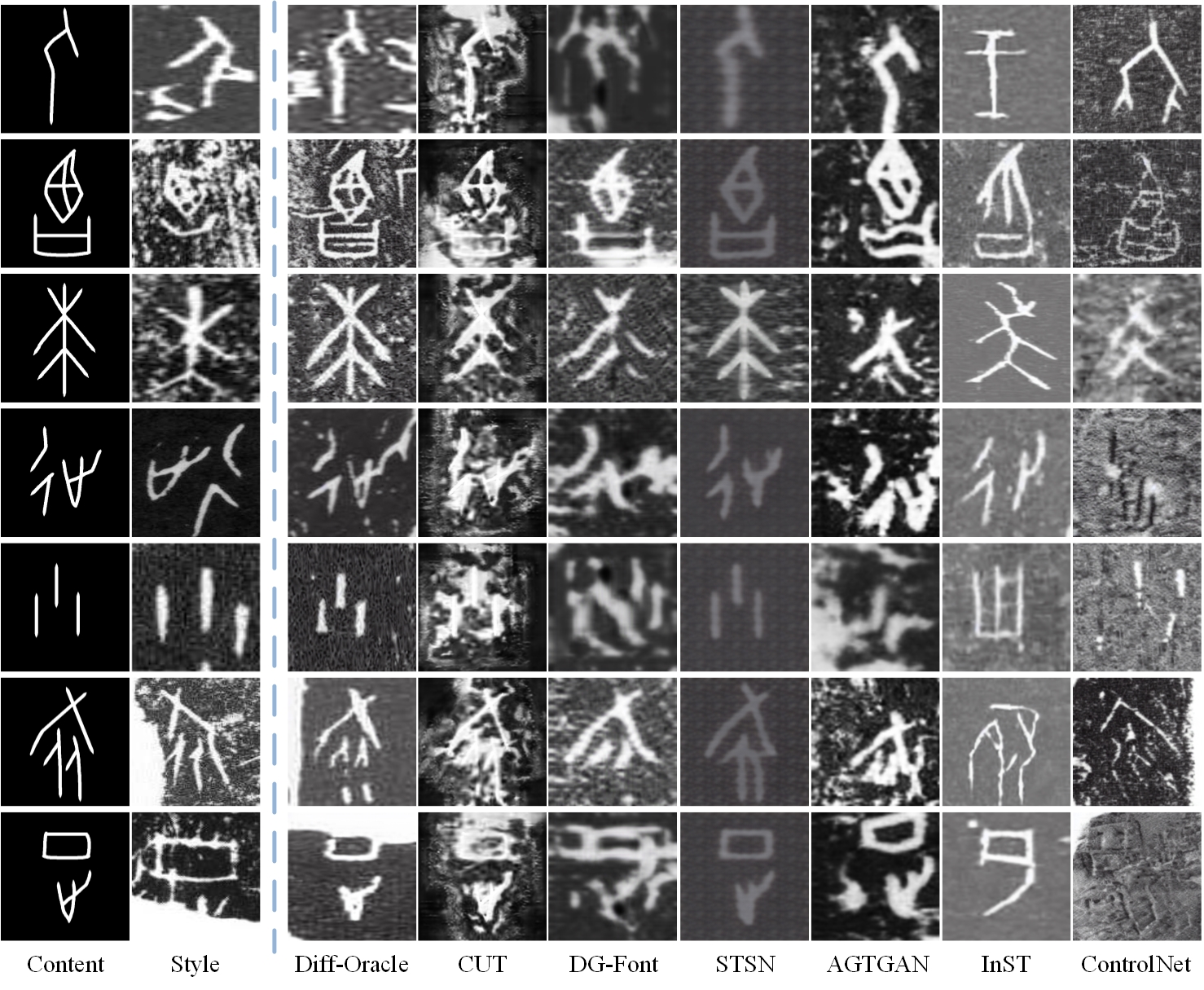}
\caption{Visualization of generated oracle character images in Oracle-241. The ``Content" column presents handprinted examples as reference content, the ``Style" column presents scanned images as reference style, and the other columns present synthesized scanned data by different generative methods.}
\label{fig:generation}
\end{figure*}

Fig.~\ref{fig:generation} shows the comparison of generation results.
It is evident that our Diff-Oracle effectively captures the stylistic attributes from scanned data, such as dense white region, missing edges, and the thickness of strokes. For instance, the reference style in the second row contains a large number of dense white regions, and our Diff-Oracle successfully incorporates this style into the generated image. Similarly, in the fourth row, we accurately reproduce the stroke thickness of the reference style while maintaining a relatively clean background akin to the original. Furthermore, in the seventh row, the character edge of the reference style is missing due to white areas; accordingly,  our final generated image also presents white areas that obscure some of the edges. Regarding content, Diff-Oracle accurately preserves the intrinsic glyphs of content images and generates meaningful character images.
In contrast, the other methods struggle to either reproduce background textures of scanned oracle characters (e.g., CUT, DG-Font and STSN) or maintain the glyphs of content images (e.g., InST and ControlNet). 
In particular, InST lacks strict content control and, therefore, cannot accurately render the glyphs of content images, leading to the generation of incorrect structures or strokes.
For ControlNet, despite utilizing real handprinted data as content images during training, it is unable to control the content of the generated images accurately. The main reason is that the real handprinted samples are paired with the scanned data only at the class level, which cannot lead ControlNet to develop the ability to precisely control content. Consequently, the resulting images often exhibit limited similarity to the reference content images or even produce meaningless characters. In addition, due to the lack of text prompts accurately describing the style of oracle characters, some generated images fail to reproduce the background texture and font style of the reference style images. 

\begin{figure}[ht]
\centering
\includegraphics[width=0.49\textwidth]{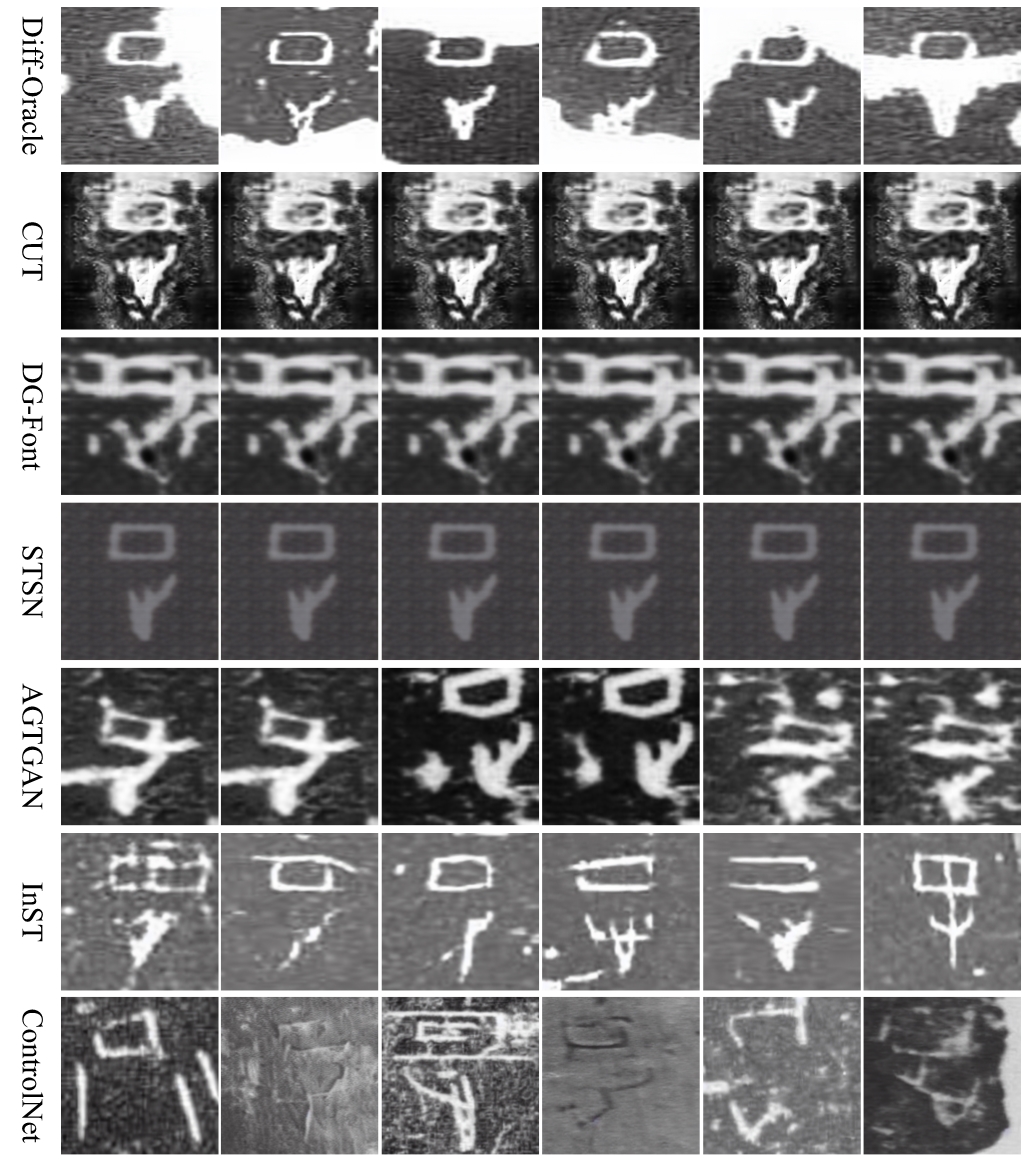}
\caption{Visualization of generated oracle character images in terms of diversity. All images belong to the same class (i.e., index ``010032" in Oracle-241) }
\label{fig:generation_div}
\end{figure}

We also compare Diff-Oracle with some other generative models in terms of diversity, as illustrated in Fig.~\ref{fig:generation_div}.  It is evident that Diff-Oracle exhibits the ability to generate diverse scanned data with consistent style and varying details (i.e., missing edges with different locations). On the contrary, many methods, such as CUT, DG-Font, and STSN, fail to produce images with high variability. AGTGAN, due to its dependence on the GAN framework, suffers from mode collapse, resulting in duplicated output types. Consequently, some generated images display duplicated styles and contents. Although InST and ControlNet, which also use the DM framework like Diff-Oracle, exhibit high diversity, the quality of their generated images is lower in terms of content and style.

\begin{figure*}[ht]
\centering
\includegraphics[width=\textwidth]{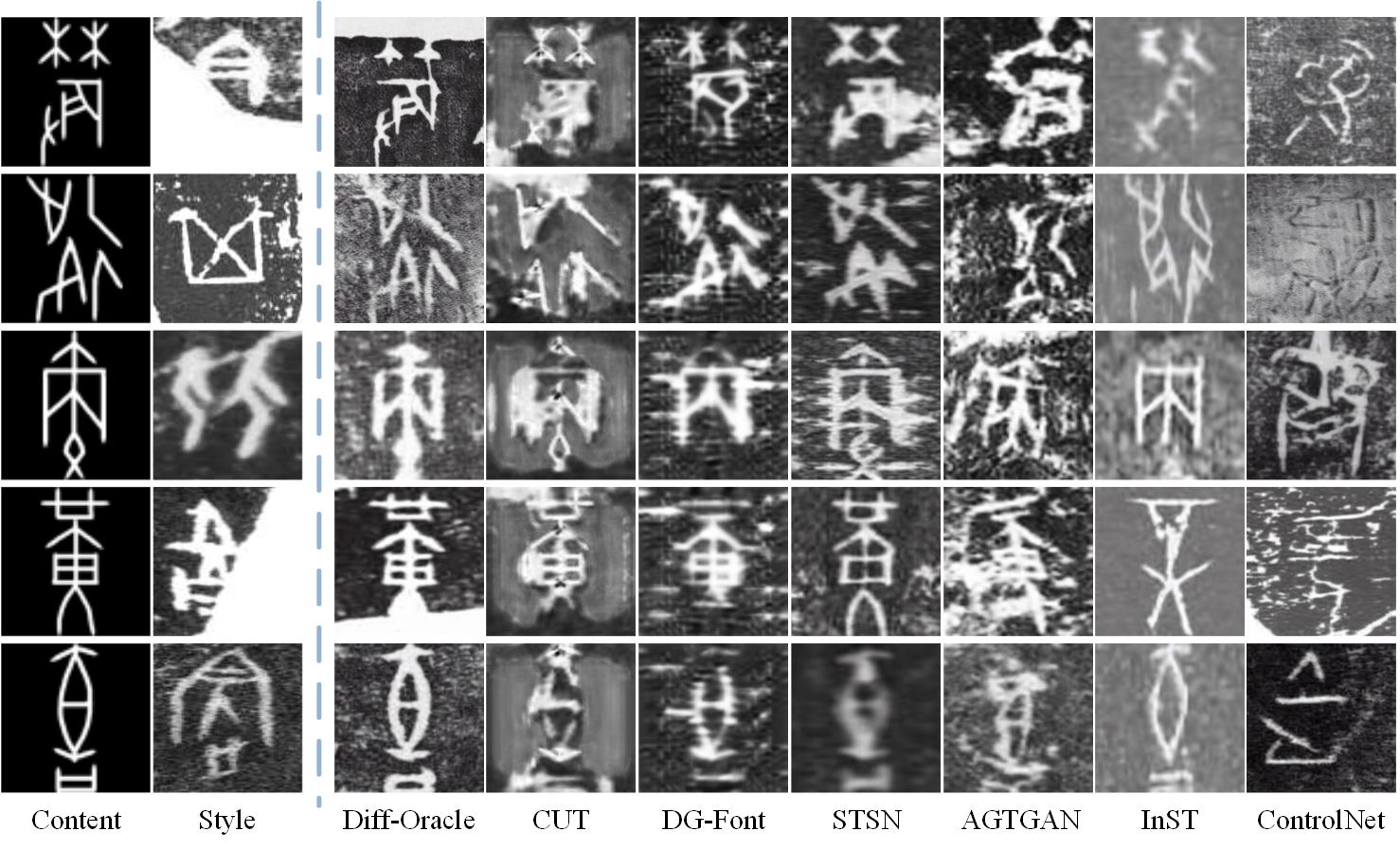}
\caption{Visualization of generated oracle character images in OBC306 for zero-shot classes. The ``Content" column presents reference contents from handprinted images in Oracle-AYNU, corresponding to zero-shot classes in OBC306. The ``Style" column represents reference styles randomly chosen from other classes in OBC306.}
\label{fig:generation_zero}
\end{figure*}

Furthermore, Diff-Oracle can generate samples for zero-shot classes. In the training set of OBC306, there are 13 classes without any training samples. Our model augments these classes by utilizing corresponding handprinted data as content and randomly selected scanned images as styles. 
For a qualitative comparison, we showcase the generated results of randomly selecting five classes in Fig.~\ref{fig:generation_zero}. We can see that our Diff-Oracle model consistently generates high-quality images with clear glyphs and realistic styles. In contrast, several methods, including DG-Font, AGTGAN, InST, and ControlNet, often fall short in generating glyphs that are both clear and precise. Moreover, some approaches generate nearly identical or duplicate instances, as observed in CUT, DG-Font, and STSN.

\subsubsection{Quantitative Comparison}
Quantitative results for generation and recognition are detailed in Table~\ref{generation_results} and Table~\ref{recognition_results}, respectively. In alignment with the previous visual analysis, consistent observations can be drawn from the tables on both datasets.

\noindent\textbf{Generation performance.}
Compared with the other methods in Table~\ref{generation_results}, it is evident that our Diff-Oracle demonstrates superior FID values of 54.1 and 57.72 in Oracle-241 and OBC306, respectively. These outcomes substantiate the fidelity of the images produced by our model. In terms of LPIPS, our model performs comparable to ControlNet. It is worth noting that, while ControlNet achieves a marginally higher LPIPS score compared to ours, the increase in diversity may be owing to the presence of chaotic and incomplete glyphs. These are deemed invalid, as illustrated in Fig.~\ref{fig:generation_div}.
\begin{table}[ht]
\centering
\caption{Comparison on Oracle-241 and OBC306 in terms of FID and LPIPS.}
\label{generation_results}
\begin{tabular}{lccccc}
\toprule
\multirow{2}{*}{Method} & \multicolumn{2}{c}{Oracle-241} && \multicolumn{2}{c}{OBC306} \\ \cline{2-3}\cline{5-6} 
            & FID$\downarrow$      & LPIPS $\uparrow$     && FID$\downarrow$     & LPIPS$\uparrow$    \\ \toprule
CUT~\cite{cut}           & 174.83       &  0.427       && 231.83      &  0.329     \\
DG-Font~\cite{2021dgfont}         & 96.50       &0.438        && 150.52      &  0.317      \\
STSN~\cite{wang2022tip}           &  208.31      &  0.335       &&83.71       &  0.476     \\
AGTGAN~\cite{huang2022acm}          & 69.32       & 0.521       && 88.7      &  0.483      \\
InST~\cite{inst}           & 86.82       & 0.479        &       &115.25&0.422       \\
ControlNet~\cite{controlnet}        &       78.66 &\textbf{0.621}        && 59.83      &   \textbf{0.648}     \\ \hline
Diff-Oracle       & \textbf{54.1}       &  0.580       && \textbf{57.72}     & 0.604      \\ \bottomrule
\end{tabular}
\end{table}

\noindent\textbf{Recognition accuracy.} We further evaluate the quality of the generated images through recognition experiments on Oracle-241 and OBC306, and the relevant results are summarized in Table~\ref{recognition_results} (as detailed in the first part of Table~\ref{recognition_results}). The proposed Diff-Oracle consistently enables the recognition model to achieve the highest accuracy in all cases for Oracle-241 and OBC306. In comparison to the baseline model without generated data~\cite{li2023towards} (the row of ``No Aug"), the average accuracy improves by 2.55\% and 8.72\% on Oracle-241 and OBC306, respectively. Notably, Table~\ref{recognition_results} shows that the performance of zero-shot classes is particularly remarkable. With the integration of generated images from Diff-Oracle, the total accuracy for zero-shot classes dramatically increases to 84.62\%. On the contrary, without content or style conditions, InST and ControlNet negatively impact both the average and total accuracy of Oracle-241 and the total accuracy of OBC306.
These results demonstrate the substantial impact of our model in enhancing the recognition accuracy, especially for zero-shot classes.

Furthermore, considering the imbalanced data problem of oracle datasets, we compare Diff-Oracle with several SOTA oracle character recognition methods~\cite{icdar2019,li2021mix,li2023towards} (as detailed in the second part of Table~\ref{recognition_results}). 
It is evident that Diff-Oracle outperforms these existing works in terms of both average and total accuracy on both Oracle-241 and OBC306. Significantly, present SOTA recognition methods depend exclusively on data from a single domain without additional auxiliary information. As a result, they are unable to handle zero-shot classes in OBC306. In contrast, Diff-Oracle demonstrates a clear advantage with an accuracy of 84.62\% on these classes.
\begin{table}[ht]
\centering
\caption{Comparison on Oracle-241 and OBC306 in terms of average, total and zero-shot class accuracy(\%). ``No Aug" means training the recognition model without using generated data. ``zero-shot" represents classes that do not have scanned data for training. The first part presents results obtained by adding generated images. The second part presents the results of oracle character recognition models. ``w/o Repatch" represents training the recognition model merely with images generated from Diff-Oracle. ``w/o generated data" represents training the recognition model merely with Repatch.}
\label{recognition_results}
\begin{tabular}{lccccc}
\toprule
\multirow{2}{*}{Method} &\multicolumn{2}{c}{Oracle-241} &\multicolumn{3}{c}{OBC306}\\ \cline{2-6} 
          & average & total  & average & total & zero-shot \\ \toprule
No Aug   & 87.92    & 89.41& 79.35    & 93.87 &-  \\
CUT~\cite{cut}    & 88.15    &  89.48 &  83.85  & 93.85 &53.85  \\
DG-Font~\cite{2021dgfont}      & 88.35    & 89.55    &     85.50 & 93.83 &    69.23	\\
STSN~\cite{wang2022tip}    &  87.92   & 89.29         &85.07 & 93.77    &    69.23	 \\
AGTGAN~\cite{huang2022acm} & 88.33    & 89.69& 85.99    & 93.82   &76.92   \\
InST~\cite{inst}     &87.40 &89.17     &84.78     &93.68 & 61.54 \\
ControlNet~\cite{controlnet}   &87.51 	&89.12 &80.05     &93.78 &15.38 \\ \hline
Zhang et al.~\cite{icdar2019} & 87.22    &88.21& 76.40    & 90.16   &-  \\
Li et al.~\cite{li2021mix}     &89.31     &90.17    &84.44     &90.67 & - \\
Li et al.~\cite{li2023towards}    &89.73     &90.27  &82.99     &93.81 &- \\ \hline

Diff-Oracle       &\textbf{90.47}   & \textbf{91.11}   & \textbf{88.07}    & \textbf{94.12} &\textbf{84.62}  \\
w/o Repatch       & 88.83   & 89.93   &   87.11  & 93.92 &\textbf{84.62}   \\
w/o generated data        & 89.13  & 90.22   & 82.44   &94.06  & -  \\
\bottomrule
\end{tabular}
\end{table}

\subsection{Ablation Studies}\label{ablation}
To showcase the effectiveness of each component in our proposed Diff-Oracle model, we conducted experiments with different components removed. 
\begin{figure}[ht]
\centering
\includegraphics[width=0.49\textwidth]{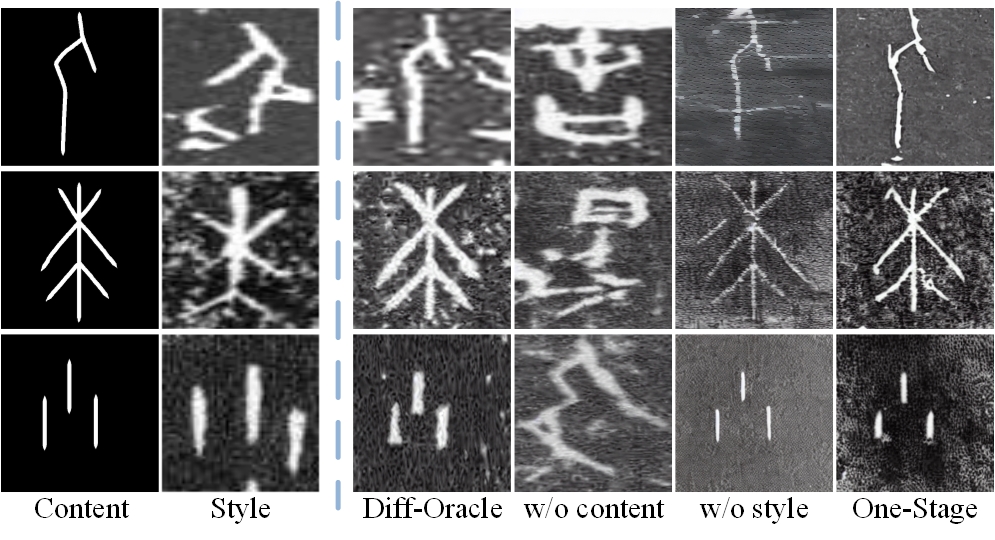}
\caption{Visualization of the effect of each component. The ``Diff-Oracle" row presents our final model, and the remaining rows present images generated under different components. ``One-Stage" represents simultaneous training content and style encoders.}
\label{fig:generation_ab}
\end{figure}
\begin{table}[ht]
\centering
\caption{Ablation study of different components of Diff-Oracle on Oracle-241. ``No Aug" means training the recognition model without generated data. ``One-Stage" represents simultaneous training content and style encoder. For recognition accuracy, Diff-Oracle here refers to the model without Repatch.}
\label{ablation_study}
\begin{tabular}{lcccc}
\toprule
\multirow{2}{*}{Method} & \multicolumn{4}{c}{Oracle-241} \\ \cline{2-5} 
            & FID$\downarrow$ & LPIPS$\uparrow$ & average$\uparrow$ & total$\uparrow$ \\ \hline
No Aug &- &- & 87.92    & 89.41   \\  
Diff-Oracle      &\textbf{54.10}   & 0.580   & \textbf{88.83}    & \textbf{89.93}   \\
w/o content       & 65.64   & \textbf{0.671}   &-  &-   \\
w/o style        & 174.77  &  0.653  & 88.17   &89.30    \\
One-Stage      & 108.86   & 0.598   &   87.88  &  88.92  \\ \bottomrule
\end{tabular}
\end{table}

\subsubsection{Effectiveness of Content Guidance} 
In the absence of content conditions (the ``w/o content" column in Fig.~\ref{fig:generation_ab}), the model is still able to produce images that possess realistic stylistic attributes like scanned data. However, it fails to control the structure of generated images, resulting in random contents that markedly diverge from reference handprinted data. This outcome stems from the fact that exclusively relying on style information cannot adequately capture the specific glyphs of oracle characters. 
In addition, if we opt not to employ pseudo content images that match pixel-level details with the scanned data, and instead directly utilize class-level paired data, the results are similar to Diff-Oracle without content. More visualization can be found in Supplementary Section~\ref{sup:unpair}. Table~\ref{ablation_study} shows that Diff-Oracle without content yields an FID value comparable to Diff-Oracle, proving that style representation controls well. Due to the chaotic content, Diff-Oracle without content obtains a higher LPIPS score. Since the random content is meaningless for recognition, we do not test the recognition performance in this case.

\subsubsection{Effectiveness of Style Guidance} 
From Fig.~\ref{fig:generation_ab}, it is evident to see that without style condition (the ``w/o style" column), Diff-Oracle generates scanned images with incongruous style and various artifacts. The main reason for this can be summarized as follows: without the guidance of style, the model does not have a clear direction for style generation. In addition, due to the discrepancy between the domains of pseudo and real handprinted data, the model trained on pseudo handprints exhibits limited generalizability to real handprint data. As a result, the quality of the generated data falls below acceptable levels. Its higher FID value in Table~\ref{ablation_study}, i.e., 174.77, further affirmed this inadequacy. Compared to the results of the baseline model (the ``No Aug" row), we can see the generated data without style control (the ``w/o style" row) does not improve the recognition performance effectively.

\subsubsection{Effectiveness of Two-Stage Training Strategy} 
We conduct an experiment to simultaneously train both the content encoder and the style encoder using the one-stage training strategy. From Fig.~\ref{fig:generation_ab}, we can see that the character images generated from one-stage training (the ``One-Stage" column) present chaotic glyphs and unrealistic styles. The possible reason is that the one-stage training prevents the model from separating style and content features effectively, which makes it challenging to handle new combinations of style and content during the generation process. As a result, the model struggles to generate the expected oracle bone script images, which in turn adversely hurt the recognition performance compared with the baseline model (the ``No Aug" row) as shown in Table~\ref{ablation_study}. These observations demonstrate the effectiveness of our proposed two-stage training strategy.

\subsubsection{Complementarity of Repatch and Generated Data}\label{repatch_ab}
To demonstrate the complementarity of Repatch and the generated data, we present the relevant results in Table~\ref{recognition_results} (as detailed in the third part of Table~\ref{recognition_results}). The results clearly demonstrate that excluding either the generated data or Repatch degrades the performance of Diff-Oracle. When these two components are combined, Diff-Oracle achieves higher performance on two datasets in terms of both average and total accuracy, highlighting the complementary nature of the data generated by Diff-Oracle and Repatch. Specifically, the generated data supplements classes with insufficient data, and also overcomes the limitation of Repatch, which is unable to support zero-shot classes. Conversely, Repatch is able to dig deeper into the training data, further improving overall recognition accuracy.

\section{Conclusion}
\label{sec:con}

In this paper, we introduce a diffusion model (DM) based framework named Diff-Oracle for generating oracle characters. To the best of our knowledge, this is the first work to explore DMs in this task. We propose two distinct encoders to separately extract style and content information, allowing for the generation of oracle characters with controllable styles and contents. For training Diff-Oracle, we generate pseudo handprinted oracle characters paired with scanned ones at the pixel level, rather than the class level. To disentangle style and content controls, we design a two-stage strategy to train both the encoders and fine-tune the diffusion model. Extensive experiments demonstrate the effectiveness of Diff-Oracle in both oracle character generation and recognition.  
Currently, we do not consider the geometric discrepancies between handprinted and scanned characters. In the future, we plan to incorporate deformable components into the generation process to further improve generation quality.

\bibliographystyle{ieeetr}
\bibliography{IEEEabrv,reference}

\clearpage
\setcounter{page}{1}
\setcounter{section}{0}
\setcounter{figure}{0}
\setcounter{table}{0}

\begin{center}
    {\Large \textbf{SUPPLEMENTARY MATERIAL}} \\
    \vspace{1.0em}
    {\Large Diff-Oracle: Deciphering Oracle Bone Scripts with Controllable Diffusion Model
    }
\end{center}

\renewcommand\thesection{\Alph{section}}
\renewcommand\thetable{S\arabic{table}}
\renewcommand\thefigure{S\arabic{figure}}

In this supplementary material, we first provide details of the datasets in Section~\ref{m_dataset}, then introduce more experimental settings in Section~\ref{scaling_factor} and Section~\ref{implementation}, and last give more experimental results in Section, Section and Section.

\section{Datasets}\label{m_dataset}

\textbf{Oracle-AYNU}~\cite{icdar2019} is a handprinted dataset, which contains 2,584 categories with 39,072 oracle character instances. Since our work focuses on scanned oracle characters, the recognition performance of this dataset is not evaluated.

\textbf{OBC306}~\cite{OBC306} is a scanned dataset, which contains 306 classes with 309,511 instances. To obtain the corresponding handprinted characters, we rely on the Oracle-AYNU dataset. Statistically, we find 275 classes in Oracle-AYNU that exactly match the characters in OBC306, while the other 31 classes are not matched. 
Consequently, in the experiments, we adopt these 275 matched classes, including 294,936 scanned samples from OBC306 and 14,799 handprinted samples from Oracle-AYNU. 
We leverage these cross-domain samples to optimize the CUT model~\cite{cut}, which can generate pseudo handprinted characters aligned with scanned characters in OBC306 at the pixel level. The generated pseudo handprinted characters are only used during the training stage of our Diff-Oracle model. Conversely, the real handprinted characters in Oracle-AYNU serve as content images to generate scanned images for OBC306 during the generation phase. 
OBC306 is randomly split into training and test sets with a 3:1 ratio for each class which follows the setting in~\cite{OBC306}. It is noted that 13 zero-shot classes lack scanned training samples, but have corresponding handprinted data.

\textbf{Oracle-241}~\cite{wang2022tip} consists of 241 classes with 78,565 samples, where each class contains both handprinted and scanned samples. It is split into the training set (10,861 handprinted samples and 50,168 scanned samples) and the test set (3,730 handprinted samples and 13,806 scanned samples). Similar to OBC306 dataset, we also leverage a CUT model~\cite{cut} to generate pseudo handprinted characters aligned with scanned characters at the pixel level, which are used in the training stage of our Diff-Oracle model.  

The statistical information of each dataset is listed in Supplementary Table~\ref{sup:table1}.

\begin{table}[htbp]
\caption{Statistical information of oracle datasets.}
\centering
\label{sup:table1}
\begin{threeparttable}
\begin{tabular}{lccccc}
\toprule
Dataset   &Type & Training Set & Test Set & Class Number \\ \hline
Oracle-AYNU &handprint & 34,424  & 4,648  & 2,584    \\
OBC306   &scan & 221,217  & 73,719  & 275   \\
Oracle-241 &scan &50,168 &13,806 &241\\
Oracle-241 &handprint &10,861 &3,730 &241\\ \bottomrule
\end{tabular}
\end{threeparttable}
\end{table}

\section{Multi-Modal Classifier-Free Guidance on Style and Content}
\label{scaling_factor}
During the generation process, we adopt a multi-modal classifier-free guidance strategy based on~\cite{goel2023pair}, which specifies different intensities for style and content controls by adjusting the corresponding scaling factors $s_1$ and $s_2$, as defined in Eq.~\ref{denoise_function}. To determine appropriate factors, we randomly select a subset of 3,000 samples from Oracle-241 and conduct a series of experiments using various scales within this subset. For convenience, we simply explore integer scales according to~\cite{diff-font,zsfontdiff}. The results are summarized in Table~\ref{guidance_scales}. 
We observe that the FID values in the first column ($s_1=0$) are much larger. This is because the generated images solely contain style information without any guidance from content. Similarly, larger FID values in the first row ($s_2=0$) correspond to inferior quality of the generated images, as they lack style guidance. By combining content and style guidance with appropriate values, Diff-Oracle can generate higher-quality oracle characters (i.e., lower FID values for all except the first row and column of Table~\ref{guidance_scales}). 
Finally, we choose $s_1=2$, $s_2=5$ for the experiments in Oracle-241. By conducting similar experiments in OBC306, we find that $s_1=2$ and $s_2=1$ are suitable scales for this dataset.
\begin{table}[ht]
\centering
\caption{Results of different scaling factors of content ($s_1$) and style ($s_2$) in Oracle-241.}
\label{guidance_scales}
\begin{tabular}{c|cccccc}
\toprule
\diagbox{$s_2$}{FID$\downarrow$}{$s_1$} &0 & 1  & 2      & 3  & 4 & 5 \\ \hline
0 &213.04	&120.39	&76.06	&80.63	&84.97	&85.73 \\
1 &204.21 & 107.80	&72.92	&75.48	&76.92	&77.67 \\
2 &183.19 & 96.70	&69.01	&70.42	&73.24	&73.05 \\
3 &162.00 & 86.35&	66.53	&68.39&	70.78	&72.05 \\
4 &152.72 & 80.58	&65.50&	67.12&	69.00&	70.46 \\
5 &137.65 & 77.24	&\textbf{64.34} &	66.29	&70.05	& 71.04 \\
6 &126.99 &72.94	&65.09	&67.03	&68.84	&71.17 \\ \bottomrule
\end{tabular}
\end{table}

\section{Experimental Setup}\label{implementation}
In the generation experiments, we set the image size to $256\times 256$ and employ the AdamW optimizer for the two-stage training of Oracle-241 and OBC306. $s_1=2$ and $s_2=5$ are used for Oracle-241, $s_1=2$ and $s_2=1$ are used for OBC306. In the first stage, the batch size is 56, and the learning rate is initially 6e-4 and decreased by 0.1 at $n_1$ within a total of $n_2$ epochs (OBC306: $n_1=25$, $n_2=40$; Oracle-241: $n_1=70$, $n_2=130$). In the second stage, the settings for the two datasets are the same. Here, the batch size is 48, and the learning rate is 6e-5 for 20 epochs. All models are trained with 4 Nvidia RTX 4090 GPUs.

In the recognition task, we adopt ViT-Base pretrained on ImageNet. SGD optimizer is adopted with momentum 0.9 and weight decay 1e-4. The learning rate is 0.01 at first, then decreased by 0.1 at the $m_1$-th and $m_2$-th epochs within total $m$ epochs (OBC306: $m_1=20$, $m_2=25$, $m=30$; Oracle-241: $m_1=40$, $m_2=50$, $m=60$). The batch size is 64, and the image size is $256 \times 256$. All models are trained on one Nvidia RTX 4090 GPU.

\section{Numbers of Generated Images}\label{number_generated}
\begin{figure}[htbp]
\centering
\includegraphics[width=0.49\textwidth]{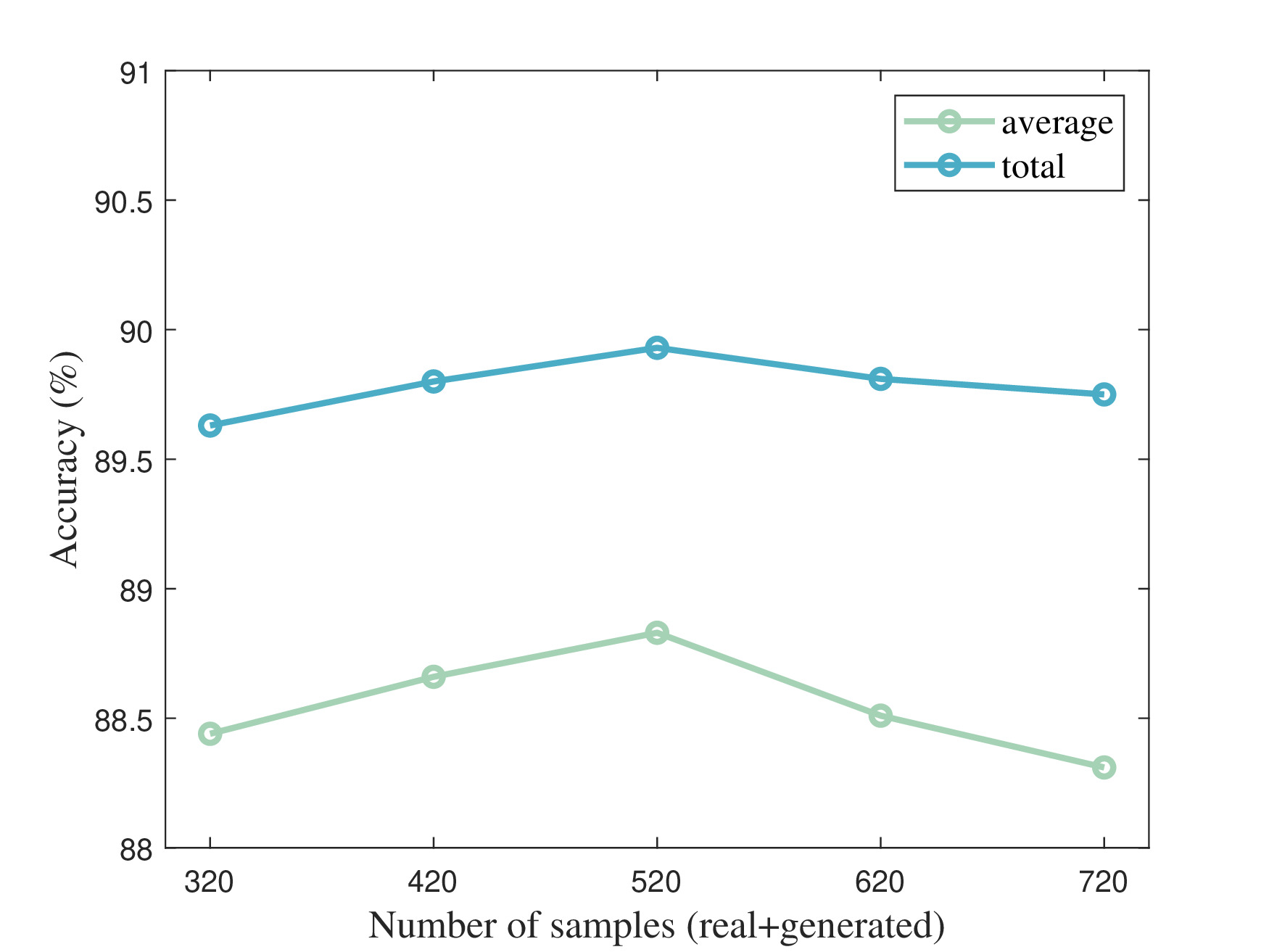}
\caption{Average and total accuracy along with the number of training samples (real and generated) in each class in Oracle-241.}
\label{fig:numbers}
\end{figure}

Diff-Oracle can generate arbitrary numbers of samples to augment the training data. To explore the impact of different numbers of generated images on recognition accuracy, we conduct the following experiments in Oracle-241. In the training set of Oracle-241, each class consists of a different number of samples, where the largest class has 320 instances. To balance all classes with the same number of samples during training the recognition model, we first generate samples to ensure that each class has 320 instances. In other words, if a class originally contains $n$ real samples, Diff-Oracle will generate $320-n$ samples for this class. Next, we increase the number of samples in each class with an increment of 100 generated samples. The recognition results are illustrated in Fig.~\ref{fig:numbers}. We can see that both average and total accuracy first increase and then decrease with the increase in the number of samples. Finally, we opt for 520 samples in our experiments. That is, Diff-Oracle generates $520-n$ new samples for each class with $n$ real samples.

As for OBC306, we simply follow the AGTGAN setup~\cite{huang2022acm}, which utilizes the generated data to augment the samples of minority classes to match the average sample size. In this paper, the average sample size of OBC306 is 805. Specifically, for classes that contain fewer than 805 training samples, we employ Diff-Oracle to generate new samples. This process ensures that each class attains a minimum of 805 samples after combining both the real and generated scanned samples. Conversely, for classes that already exceed 805 samples, no additional generated images are added to train the recognition model.

\section{Effectiveness of Pixel-Level Paired Samples}
\label{sup:unpair}
\begin{figure*}[htbp]
\centering
\includegraphics[width=0.7\textwidth]{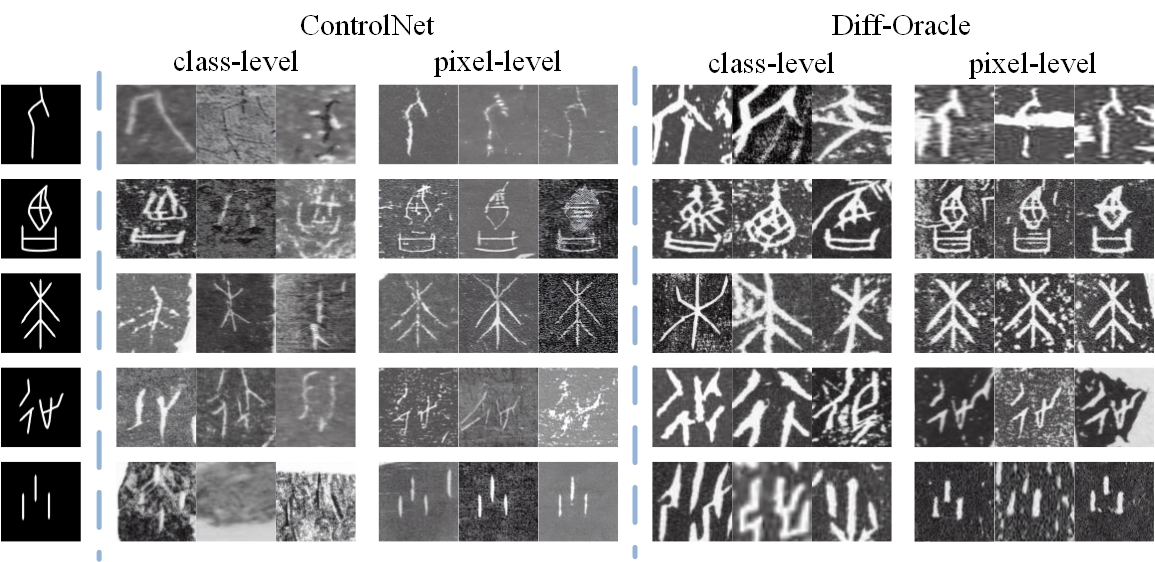}
\caption{Comparison of generated oracle character images in Oracle-241 by models trained with class-level paired data or pixel-level paired data.}
\label{fig:generation_unpair}
\end{figure*}

To demonstrate the importance of pixel-level paired data, we conduct experiments using class-level pairs to train ControlNet and Diff-Oracle, and the generated samples are shown in Fig.~\ref{fig:generation_unpair}.
Trained on class-level paired data, ControlNet struggles to produce accurate structures, such as incorrect glyphs or inserting extra strokes. Although this model integrates content images, its control over the content structure is limited due to the coarse match between content images and input images during training.
Conversely, when trained with pixel-level paired data, ControlNet generates more precise glyphs due to the precise guidance at the pixel level. 
Similarly, Diff-Oracle trained with class-level paired data exhibits structures inconsistent with content images, despite showing some degree of similarity. After changing to pixel-level paired data, Diff-Oracle is able to accurately control the structure of generated images. 
In summary, the pixel-level paired data can significantly enhance the content control of the generated results, thereby improving the generation quality.

\end{document}